\documentclass[11pt]{article}

\usepackage[final]{acl}

\makeatletter
\@ifpackageloaded{lineno}{%
  \AtBeginDocument{%
    \def\makeLineNumber{%
      \if@firstcolumn\makeLineNumberLeft\else\makeLineNumberRight\fi}%

  }%
}{}
\makeatother

\usepackage{times}
\usepackage{latexsym}

\usepackage[T1]{fontenc}

\usepackage[utf8]{inputenc}

\usepackage{microtype}

\usepackage{inconsolata}

\usepackage{graphicx}

\usepackage{xcolor}

\usepackage{listings}
\lstdefinestyle{jsonstyle}{
  basicstyle=\small\ttfamily,
  breaklines=true,
  breakatwhitespace=false,
  columns=fullflexible,
  showstringspaces=false,
  keepspaces=true,
  upquote=true,
}

\usepackage{booktabs, multirow} 
\usepackage{soul}
\usepackage{xcolor,colortbl} 
\usepackage{changepage,threeparttable} 
\usepackage{amsmath}
\usepackage{float}

\usepackage{CJKutf8} 
\usepackage{ragged2e}
\usepackage{enumitem}
\title{MortarBench: Evaluating Mortgage Loan Origination Agents}

\author{
 \textbf{Matthew Toles\textsuperscript{1}},
 \textbf{Yunan Lu\textsuperscript{1}},
 \textbf{Manav Munjal\textsuperscript{1}},
 \textbf{Bojun Liu\textsuperscript{1}},
\\
 \textbf{Yuanhao Deng}\textsuperscript{2},
 \textbf{Stephanie Selig\textsuperscript{2}},
 \textbf{Derek Rindner\textsuperscript{2}},
 \textbf{Cheng Li\textsuperscript{2}},
 \textbf{Zhou Yu\textsuperscript{1}}
\\
\\
 \textsuperscript{1}Columbia University,
 \textsuperscript{2}Tidalwave
\\
 \small{
   \textbf{Correspondence:} \href{mailto:mt3639@columbia.edu}{mt3639@columbia.edu}
 }
}

\begin{document}
\maketitle
\begin{abstract}

Loan origination is the process by which a lender creates a new loan, from application and underwriting through approval and funding.
This process serves a critical role in evaluating the eligibility and level of risk posed by an applicant.
Recently, firms have begun using mortgage loan agents to augment human loan officers, despite a lack of any public benchmark.
To fill this gap, we present MortarBench, a loan origination agent benchmark.
MortarBench uses a financial data synthesis and mutation pipeline to generate examples with broad edge case coverage that match real-world distributions and questions.
We find that state-of-the-art large language models (LLMs) perform poorly, with closed-source models achieving at most 80.8\% F1 accuracy.
We also discover systematic biases in LLM perception of foreignness related to non-English names.
Noting these weaknesses, we introduce CRIT, a confidence calibration framework.
Our method increases peak F1 accuracy to 83.6\% while improving risk management steering and reducing bias.

\end{abstract}

\section{Introduction}

The US mortgage industry generates over \$1.7 trillion through loan origination annually, necessitating a complex web of risk assessment and regulatory compliance \cite{mba2024forecast}. 
Mistakes during these decisions carry severe legal and financial consequences, including loan defaults, liabilities, and fines for non-compliance.
Although manual review is costly and automation is appealing, fully replacing human oversight introduces significant risk.

\begin{figure}
    \centering
    \includegraphics[width=1.0\linewidth]{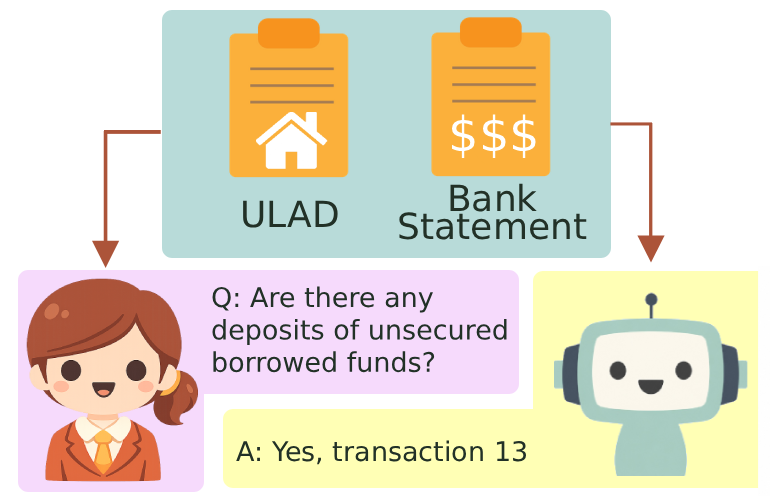}
    \caption{Example MortarBench task. A mortgage underwriter is provided with a bank statement and ULAD loan applicant financial profile. The underwriter asks the chatbot assistant a question about the documents, which it answers based on those same documents.}
    \label{fig:main}
\end{figure}

The loan origination process focuses on validating specific facts about an applicant's financial status (Figure \ref{fig:main}) against their claims according to guidelines set by governing bodies such as Fannie Mae and Freddie Mac. 
Underwriters primarily rely on 1) standardized documents containing the borrower's self-reported profile (assets, liabilities, accounts, marital status, etc.), and 2) official bank statements. 
The officer's goal is to ensure consistency between the profile and bank statements, and that the application adheres to regulatory guidelines.
However, analyzing the application package, and especially bank statements, is a difficult and tedious process.
Because bank statements and transaction descriptions do not follow any standard format, reconciliation for a question such as ``What is the total value of the applicant's unsecured loans?'' is in fact a complex natural language and numerical task.

Recently, loan origination firms have begun integrating LLM agents into their workflows.
Despite this rapid integration, the industry lacks a standardized benchmark to evaluate these agents' performance.
This evaluative vacuum hinders broader adoption, stifles iterative development, and prevents objective comparisons between competing products.
To address this gap, we propose MortarBench, a novel evaluation framework designed specifically for mortgage underwriting tasks.

Generating realistic, challenging examples of expert tasks in MortarBench presents two major challenges.
First, real financial data generally cannot be shared publicly due to privacy concerns, so we must generate internally consistent synthetic data that matches real-world distributions.
Second, we must find a way to steer synthetic examples to cover rare edge cases relevant to user questions.
We overcome both challenges using a mutation-based synthetic data generation pipeline that allows us to procedurally edit input documents to force a desired answer while maintaining internal consistency across application documents.

Finding that baseline foundation models achieve weak accuracy on MortarBench, we contribute the CRIT Agent, a confidence-calibrated model that improves accuracy, bias, and steerability.
CRIT improves accuracy on 6/7 baseline models, achieving a 2.8\% gain (14.6\% error reduction) in Gemini 3.1 Pro for a state-of-the-art 83.6\% F1 accuracy.

\section{Related Work}
Prior work has explored financial regulatory question answering (QA), including \citet{sohn2021global} and \citet{chen2024fintextqa}.
These works focus on public-facing questions (e.g. FAQ, textbooks) rather than private, in-production systems.
Work on financial numeric and table QA includes \citet{zhu2021tat, chen2021finqa, reddy2024docfinqa, islam2023financebench}.
\citet{trivedi2024tabsniper} introduces a dataset for bank statement table detection and structure recognition but without a QA component.
\citet{choi2025finder} analyzes retrieval-augmented generation (RAG) QA.
\citet{loukas2022finer} studies entity recognition in financial reports.
\citet{mollaev2025multimodal} releases large-scale anonymized transaction, geo-position, and technical support chat data from a major bank.
In contrast, we contribute a benchmark in the mortgage origination domain focused on the chatbot assistant role.
To our knowledge, this is the first public dataset for QA directly on bank statement natural language and numerical contents.
Numerous works have explored the ability of LLMs to self-calibrate confidence in their own answers, including \citet{manakul2023selfcheckgpt, zhu2023calibration, kapoor2024calibration, kuhn2023semantic, kadavath2022language}.
In particular, we draw on \citeposs{zhu2023calibration} recognition that large models can accurately estimate their own confidence.

\makeatletter
\ifacl@anonymize
We will publicly release our code and data.
\else
We release our code and data at \href{https://github.com/mtoles/MortarBench}{https://github.com/mtoles/MortarBench}
\fi
\makeatother


\begin{figure}[h]
    \centering
    \includegraphics[width=0.5\textwidth]{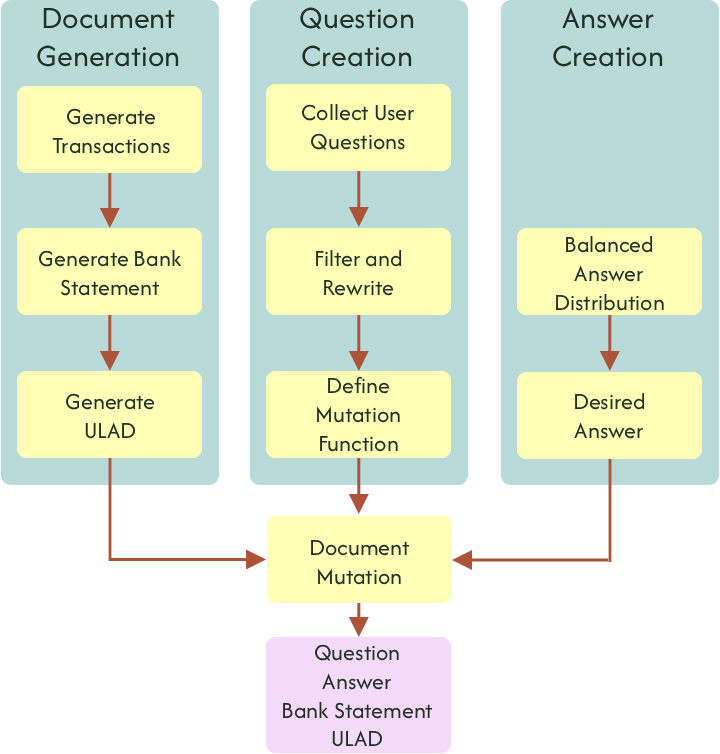}
    \caption{Dataset generation pipeline. We generate transactions across one or more bank statements to match real-world distributions, then create an applicant profile (ULAD) that conforms thereto. We filter and rewrite questions derived from users of an in-production chatbot assistant. For each question, we manually create a mutation function that edits the transaction and/or ULAD to conform to a desired answer.}
    \label{fig:pipeline}
\end{figure}

\section{MortarBench Creation Process}

To address the lack of standardized evaluation in the mortgage industry, we introduce MortarBench --- \textbf{Mor}tgage \textbf{T}ransaction \textbf{A}nalysis and \textbf{R}easoning Benchmark --- a specialized benchmark for evaluating LLM agents on realistic questions from loan officers.

Each data instance in MortarBench (Figure \ref{fig:pipeline}) consists of the following components:

\noindent \textbf{Bank Statement:} A JSON-formatted list of financial transactions spanning 60 days.

\noindent \textbf{ULAD File:} The Uniform Loan Application Dataset (ULAD), an industry-wide standardized form containing a summary of the applicant's financial profile (e.g., employment history, stated assets, and liabilities).

\noindent \textbf{Target Question:} A single question about the bank statement and, optionally, the ULAD, based on real-world loan officer workflows, curated and refined by expert annotators.

\noindent \textbf{Ground-Truth Answer:} The answer corresponding to the question, expressed as either a Boolean value, a list of transaction identifiers, or a list of account identifiers.

\subsection{Bank Statement Generation}
\label{sec:bank-statement}

To generate realistic transaction data that captures complex financial activity and supports diverse underwriting questions, we construct bank statements grounded in real financial records.
We sample 300 real bank statements from real loan applicants processed by a medium-sized loan processing company.
To overcome privacy limitations, we extract anonymous summary statistics of these applications, including the frequency and average value of key transaction types: payroll, benefits, and other income; deposits; housing and utilities expenses; bank fees; cryptocurrency transactions; and buy now, pay later (BNPL) and other liabilities.
These statistics are used to generate synthetic bank statements with matching distributions.
Finally, we manually tag each transaction with transaction properties such as \texttt{unsecured loan} based on industry definitions.
Tags will be used later to calculate profile information and force specific answers in examples.
We limit bank statements to 60 days, aligning with regulatory guidelines.
We present an example bank statement in Appendix \ref{sec:app:bank-statement}.

\subsection{ULAD Generation}
\label{sec:ULAD-gen}

To generate ULAD documents consistent with the bank statement, we adopt a template-driven approach based on the standardized ULAD schema \citep{freddiemac_ulad}.
We populate each field relevant to dataset questions, including assets, collateral, liabilities, and loans with values computed from the bank statement using transaction property tags.
We present a summary of ULAD contents in Appendix \ref{sec:app:ulad}.





\subsection{Question Creation}
\label{sec:question-creation}

To generate realistic user questions, we collect real-world user questions made to an in-production loan assistant chatbot.
We manually filter these questions to those that are 1) answerable using logical and financial reasoning or the Fannie Mae underwriting guide \citep{fanniemae_selling_2025} and 2) prevalent and useful within the loan origination process according to two subject matter expert authors. 
Authors then manually rewrite these questions to avoid multi-turn and temporal dependencies; ambiguous user intent; subjective questions; and questions that could leak personally identifiable information.

We find that answers fall into three categories:

\noindent \textbf{Boolean} - Answered with yes or no. Example: Do the payroll deposit entries match with the primary borrower's employer names stated in employment history?

\noindent \textbf{Transaction List} - Answered with a list of transaction IDs. Example: Which list of deposits, if any, are considered as large deposits?

\noindent \textbf{Account List} - Answered with a list of account IDs. Example: Which list of accounts, if any, are joint accounts where one or more account holders are not listed as borrowers on the loan application?

We also identify exactly one question requiring a monetary value answer.

\subsection{Answer Creation and Document Mutation}

To generate balanced benchmark instances at scale, we design mutation functions that modify ULAD and bank statement files according to target questions and desired answers.
For example, for the question ``How many buy now, pay later (BNPL) transactions occur in the transaction list?'' with the desired answer ``3'', the mutation adds or removes existing BNPL transactions using the transaction tags defined in Section \ref{sec:bank-statement} until there are exactly three BNPL transactions.
If necessary, we update values in the ULAD, such as total liabilities, to reflect changes in the bank statement.
Unlike with fully random bank statement generation, this process creates a balanced dataset while preserving inter-document consistency.

\subsection{Summary Statistics}

We identify 47 unique questions satisfying the criteria in Section~\ref{sec:question-creation}.
For each question, we generate 4 cases, each with a profile consisting of a ULAD and a bank statement.
For Boolean questions, we mutate half of the profiles to yield a ``yes'' answer and the other half to yield a ``no'' answer.
For transaction and account list questions, we mutate half of the profiles to contain a random non-zero number of transactions or accounts, and the other half to contain none.

\begin{table}[!htp]\centering\small
\begin{tabular}{lrr}\toprule
Total test cases &188 \\
Number of unique questions &47 \\
Avg.\ transactions per bank statement &33.6 \\
ULAD fields &162 \\
Questions needing ULAD &53.2\% \\
Boolean answers &29.8\% \\
Transaction list answers &57.4\% \\
Account list answers &10.6\% \\
\bottomrule
\end{tabular}
\caption{MortarBench Dataset Statistics}\label{tab:mortarbench-stats}
\end{table}

\subsection{Evaluation Metrics}
We evaluate agents based on exact match (EM) and F1 to allocate partial credit to list-type questions.
For Boolean questions, we assign an F1 score of $1$ for correct predictions and $0$ otherwise.

\section{Baseline Results}

We find that Gemini 3.1 Pro and DeepSeek V4 Pro achieve the greatest F1 and exact match accuracy, respectively (Table \ref{tab:eval_mt_combined_nostd}). 
They differ significantly in their strengths with respect to question type; Gemini performs well on Boolean and Account ID questions while DeepSeek performs better on Transaction ID questions.
Overall, we see the weakest accuracy across models in the Transaction ID category, with a maximum F1 of 79.7\%.
All models produce far more false positives compared to false negatives on Transaction ID questions.
This may be due to the relatively large number of true negatives compared to true positives in Transaction ID type questions, as well as models' tendency to guess rather than admit they cannot answer \cite{kalai2026evaluating}.

\section{Methods}

Noting weak baseline model performance, especially on Transaction ID questions, we introduce the \textbf{C}onfidence \textbf{R}eflection \textbf{I}nference for \textbf{T}ransactions (CRIT) Agent.
CRIT is motivated by the severe imbalance between questions with false positives per trial (36.7) vs. questions with false negatives (0.0) in Gemini and baseline models, indicating oversensitivity.
CRIT differs from the baseline implementation in that, for each transaction in its answer, the agent also generates a confidence score (1-5) indicating the probability the transaction is correctly included.
The confidence is based on the number and strength of assumptions that must be made to justify the inclusion.
Finally, we drop all transactions with confidence below a threshold $T$.
We choose $T=5$ based on a parameter sweep using Gemini (Figure \ref{fig:fp_vs_fn}), but hold $T=5$ constant for all models.
Confidence adjustment serves both to align the model away from oversensitivity and to allow steering depending on the relative cost of false positives vs negatives.
Full prompt templates for implementations are provided in Appendix \ref{appendix:prompts}.


\section{Experimental Results}

Our results show that overall, Gemini 3.1 Pro performs strongest on MortarBench, both with and without the CRIT framework.
CRIT improves 6/7 base models' F1 and EM (Table \ref{tab:eval_mt_combined_nostd}).
The gain is largest for the Kimi and Qwen baselines (8.6\%).
CRIT does not improve accuracy on DeepSeek V4 Pro. 
Although CRIT reduces false positives as expected, the remaining false positives are distributed across a wider range of questions.
This results in an overall increase in questions containing at least one false positive and therefore a decrease in overall accuracy. 

We additionally attempted to improve models' domain knowledge by adding a RAG module to the baseline pipeline.
We included the Fannie Mae selling guide in the RAG database but found that model accuracy either stayed the same or degraded.
We believe this is because the type of domain knowledge present in the selling guide differs from what is needed here.
The selling guide is dominated by procedural matters (e.g. lender breach of contract and record keeping) rather than how to handle contextual edge cases (e.g. whether a monthly Venmo payment to a roommate qualifies as a rent payment).

\begin{table*}[t]
\centering
\small
\resizebox{\textwidth}{!}{%
\begin{tabular}{lrrrrrrrrrr}
\toprule
 & \multicolumn{2}{c}{Overall} & \multicolumn{3}{c}{F1 By Question Type} & \multicolumn{2}{c}{Extras (FP)} & \multicolumn{2}{c}{Drops (FN)} & \\
\cmidrule(lr){2-3} \cmidrule(lr){4-6} \cmidrule(lr){7-8} \cmidrule(lr){9-10}
Method & F1 & EM & Boolean & Txn IDs & Account IDs & Qs & FP/Q & Qs & FN/Q & $p$ \\
\midrule
Claude Sonnet 4.6 & 56.1\,{\scriptsize(54.6--56.9)} & 51.4\,{\scriptsize(50.5--52.1)} & 36.3 & 63.0 & 85.0 & 48.7 & 2.5 & 1.7 & 0.05 & -- \\
\quad + CRIT & 61.6\,{\scriptsize(60.2--62.9)} & 58.7\,{\scriptsize(57.1--59.8)} & -- & 72.6 & -- & 34.3 & 1.2 & 2.7 & 0.05 & \textbf{$<$0.001} \\
\midrule
GPT-5.5 & 79.9\,{\scriptsize(79.2--80.8)} & 76.8\,{\scriptsize(76.1--77.1)} & 94.6 & 72.3 & 93.3 & 34.7 & 1.9 & 1.3 & 0.03 & -- \\
\quad + CRIT & 80.4\,{\scriptsize(79.4--81.1)} & 77.5\,{\scriptsize(77.0--78.0)} & -- & 73.1 & -- & 31.0 & 1.2 & 7.7 & 0.14 & 0.31 \\
\midrule
Gemini 3.1 Pro & 80.8\,{\scriptsize(80.0--82.2)} & 77.1\,{\scriptsize(76.1--78.2)} & \textbf{97.0} & 72.2 & \textbf{96.7} & 36.7 & 1.1 & \textbf{0.0} & \textbf{0.00} & -- \\
\quad + CRIT & \textbf{83.6}\,{\scriptsize(83.1--84.3)} & \textbf{80.5}\,{\scriptsize(80.1--81.2)} & -- & 77.0 & -- & 29.3 & 1.0 & 3.3 & 0.06 & \textbf{$<$0.001} \\
\midrule
DeepSeek V4 Pro & 80.7\,{\scriptsize(78.8--82.0)} & 78.0\,{\scriptsize(76.1--79.3)} & 89.3 & \textbf{79.7} & 76.1 & \textbf{25.7} & 1.9 & 0.7 & 0.01 & -- \\
\quad + CRIT & 77.8\,{\scriptsize(77.3--78.1)} & 75.2\,{\scriptsize(75.0--75.5)} & -- & 74.6 & -- & 29.7 & 1.6 & 5.0 & 0.10 & 1.00 \\
\midrule
Kimi K2.6 & 64.0\,{\scriptsize(62.7--65.5)} & 60.5\,{\scriptsize(59.0--61.2)} & 66.1 & 63.4 & 74.0 & 42.3 & 3.2 & 5.7 & 0.12 & -- \\
\quad + CRIT & 72.6\,{\scriptsize(72.2--73.1)} & 70.0\,{\scriptsize(69.5--70.6)} & -- & 78.5 & -- & 26.7 & 1.1 & 5.3 & 0.09 & \textbf{$<$0.001} \\
\midrule
Qwen3 32B & 51.2\,{\scriptsize(48.8--54.6)} & 46.5\,{\scriptsize(44.7--48.9)} & 42.3 & 54.8 & 66.9 & 54.0 & 2.0 & 11.0 & 0.17 & -- \\
\quad + CRIT & 59.8\,{\scriptsize(59.0--60.5)} & 56.7\,{\scriptsize(55.7--57.8)} & -- & 69.8 & -- & 29.7 & \textbf{0.5} & 14.0 & 0.23 & \textbf{$<$0.001} \\
\midrule
Mistral Small 24B & 38.5\,{\scriptsize(30.9--43.2)} & 32.3\,{\scriptsize(26.1--36.2)} & 35.7 & 41.9 & 35.8 & 72.3 & 2.1 & 18.0 & 0.27 & -- \\
\quad + CRIT & 42.4\,{\scriptsize(40.1--45.3)} & 38.3\,{\scriptsize(36.5--41.3)} & -- & 48.6 & -- & 44.3 & 1.0 & 30.7 & 0.63 & \textbf{0.01} \\
\bottomrule
\end{tabular}%
}
\caption{Accuracy overall (F1, EM) and by question type (F1 \%, mean over $N=3$ trials), alongside error counts on transaction-list questions: the share of questions (Qs) with at least one false positive (FP) or false negative (FN) transaction and the average number of FP and FN transactions per question. CRIT edits only transaction lists, so its Boolean and Account ID scores are identical to the corresponding baseline and are shown as --.}
\label{tab:eval_mt_combined_nostd}
\end{table*}

\section{Failure Analysis and Discussion}

Current frontier models demonstrate diverse strengths and weaknesses across subtasks. 
Baseline Gemini 3.1 Pro achieves state-of-the-art performance on Boolean and Account ID questions but places third after DeepSeek V4 Pro and GPT-5.5 on Transaction ID questions.
However, these models perform worse on Boolean and Account ID questions.
Performance on transaction list questions is substantially weaker, indicating a continued need for human-in-the-loop oversight.

\subsection{Failure Analysis}

To identify reasoning failures, authors manually reviewed all 36 erroneous reasoning traces in baseline Gemini on Transaction ID questions, grouping errors qualitatively.
We choose Gemini because it demonstrated the strongest overall baseline F1 accuracy, and thus the most likely choice for industry practitioners to build on.
We identify four distinct types of errors.
The largest portion of errors (transaction misclassification, 33.3\%) occur due to erroneously adding or omitting a transaction based on its description, for example, including a personal loan as a BNPL transaction.
27.8\% of errors occur due to failed value matching, i.e., Gemini assumes that a set of rent payments \textbf{exceeding} the total stated in the ULAD should be considered valid, when it is not.
Domain knowledge errors (22.2\%) include falsely assuming that all wire transfers are international or that deposits by co-borrowers are automatically documented.
In 11.1\% of cases, Gemini misinterprets the constraints in the prompt, for example, classifying a one-time housing payment as recurring because it might recur in the future.
We expect that value matching and constraint misinterpretation errors may be improved by increasing the specificity of prompts, i.e., that the LLM lacks sufficient understanding of industry norms to accurately interpret user intent.
Transaction classification and domain knowledge errors, however, likely require an additional knowledge base or models with better parameterized knowledge.

\begin{table}[H]
\centering
\small
\begin{tabular}{lrr}
\toprule
Category & Count & \% of Total \\
\midrule
Transaction Misclassification & 12 & 33.3 \\
Value Matching Error & 10 & 27.8 \\
Lack of Domain Knowledge & 8 & 22.2 \\
Constraint Misinterpretation & 4 & 11.1 \\
Other Error & 2 & 5.6 \\
\bottomrule
\end{tabular}
\caption{Failure-mode breakdown over $N=36$ wrong predictions on transaction-list questions for Gemini.}
\label{tab:failure_modes_txn}
\end{table}

\begin{figure}
    \centering
    \includegraphics[width=0.99\linewidth]{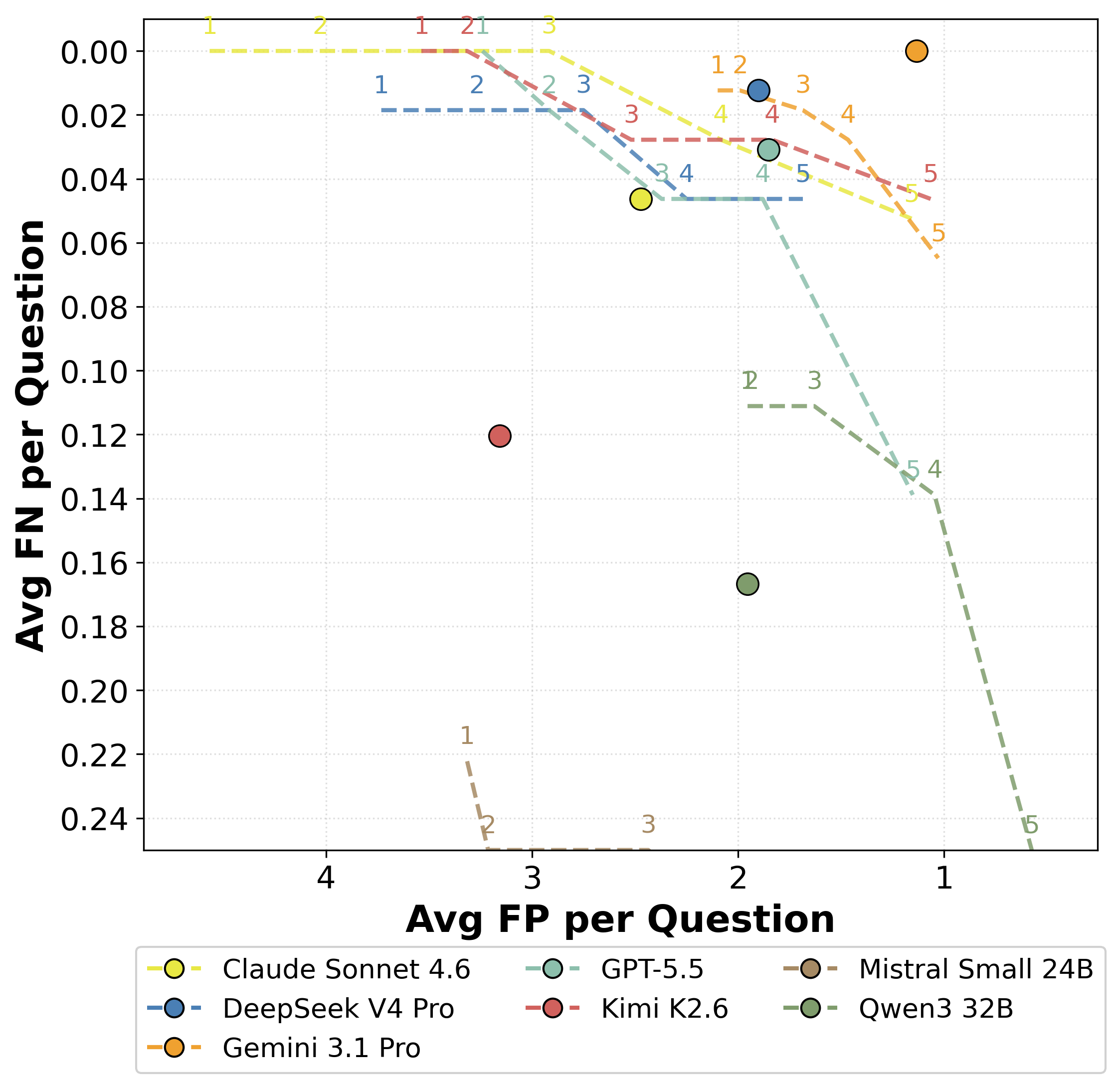}
    \caption{False negative and positive rate as a function of CRIT threshold $T$. Baseline results indicated with circles. Note that axis scales are chosen to ensure that small changes in FN are visible. Due to high FN rate, Mistral is partially clipped.}
    \label{fig:fp_vs_fn}
\end{figure}

CRIT successfully aligns base models to reduce oversensitivity in transaction selection, reducing the false positive rate significantly at the expense of a negligible increase in false negatives (Table \ref{tab:eval_mt_combined_nostd}).
In addition to improving overall EM and F1 compared to baseline, selective thresholding also improves steering and interpretability (Figure \ref{fig:fp_vs_fn}).
It can be adjusted to reflect real-world financial risk or human verification cost associated with each error.
Additionally, self-reflective confidence values can be used by users to prioritize human review.

\subsection{Stochasticity Analysis}
\label{sec:stochasticity}


Model failure typically follows one of two consistency patterns. 
Given a challenging problem, the model either guesses randomly (stochasticity), or answers consistently incorrectly based on spurious reasoning (consistent false belief).
By generating two document sets for each question-answer pair, we can systematically measure model failure mode consistency.
We quantify the consistency of errors using risk ratio (RR) \citep{rothman2008modern}, defined as how much a first-version error raises the probability of a second-version error:

\begin{align*}
\mathrm{RR} = \frac{\Pr(\text{2nd wrong}\mid\text{1st wrong})}{\Pr(\text{2nd wrong}\mid\text{1st right})}
\end{align*}

\noindent An RR of 1 would indicate that the two errors are independent—pure stochasticity. For baseline Gemini, however, an error on the first version makes an error on the second 10.7× more likely than a first-version success does (3 trials, 129 total errors). 
This is the greatest risk ratio among models. 
Other models express RR ranging from 3.0 (Mistral) to 6.8 (GPT-5.5) (Appendix~\ref{appendix:stochasticity}).
We therefore conclude that the overwhelming majority of errors stem from false beliefs in the model rather than from stochastic variation. 
This suggests models could benefit from additional in-context domain expertise and relevant post-training.

\subsection{Bias Analysis}

\begin{figure*}
    \centering
    \includegraphics[width=1\linewidth]{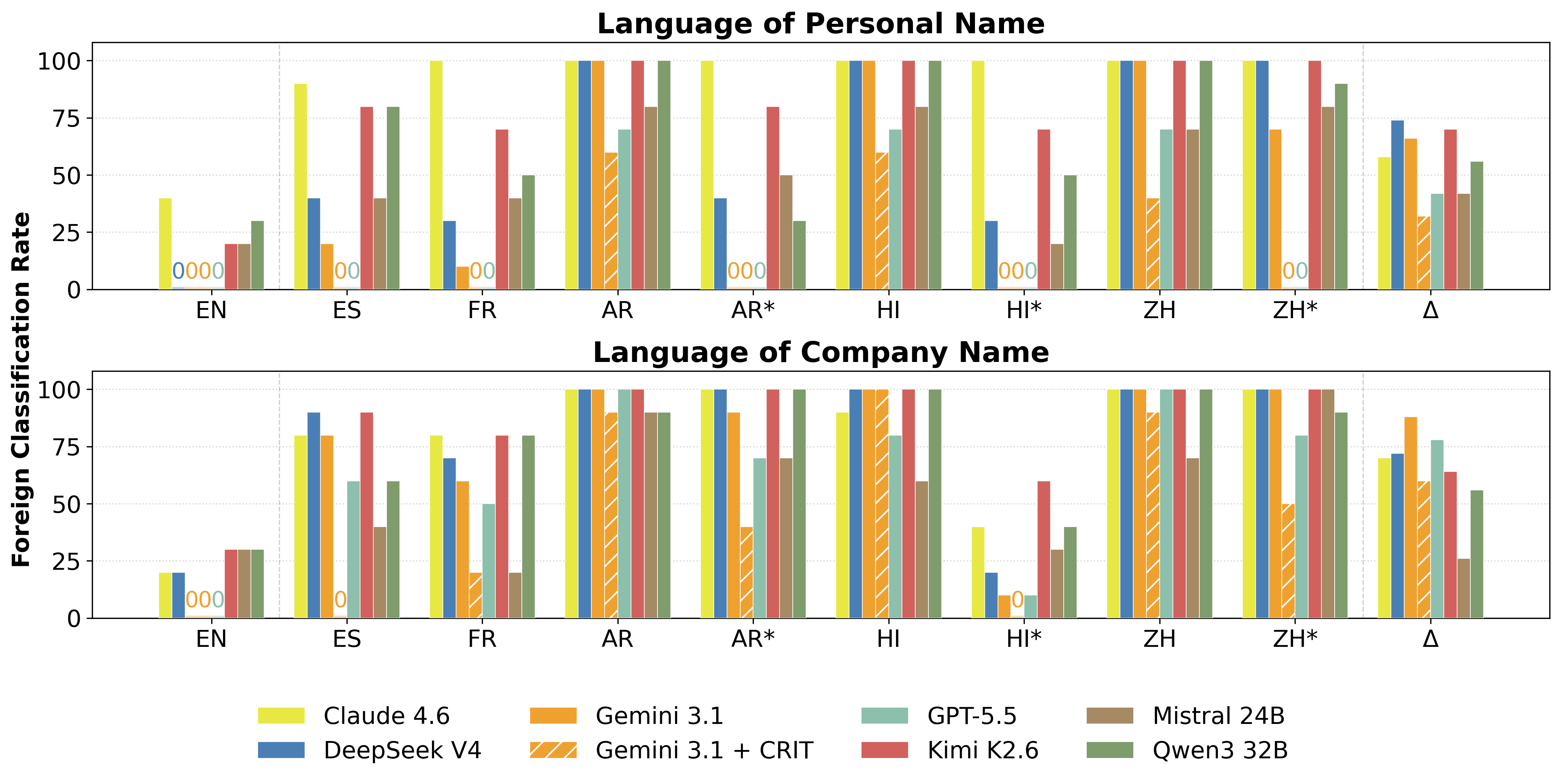}
    \caption{Rate at which wire transfers are classified as foreign as a function of language for company and personal names. * indicates that names have been transliterated into Latin script. $\Delta$ indicates the difference between the mean of all non-English languages in their native script and English.}
    \label{fig:name_bias}
\end{figure*}
For legal and ethical reasons, it is important that lending institutions do not discriminate against applicants based on protected classes.
Through manual review, we identify two scenarios with elevated potential for biased decisions: 
identifying foreign-origin transactions and classifying transactions involving informal value transfer systems (IVTSs).

We probe bias by inserting transactions with the description ``ACH CREDIT - \{entity name\}'' into the bank statement and measuring whether the retrieval rates differ between the control and experimental groups.

\subsubsection{Bias Involving Non-English Names}
To test whether the LLM perceives non-English names as more foreign, we set \{entity name\} as either an English or a non-English personal or company name and prompt the model with ``Which list of deposits, if any, could be of foreign origin?''.
For personal names, we use the five most common masculine and feminine given names and the ten most common family names, paired arbitrarily, from Wikipedia's lists of most common given names and surnames \citep{wiki_given_names, wiki_surnames}.

We observe an extreme level of bias based on the language of origin of the name.
Across all seven baseline models, transfers from English names are classified as foreign origin in only 15.7\% of cases (Figure \ref{fig:name_bias}).
Gemini, DeepSeek, and GPT-5.5 never classify these transactions as foreign origin.
However, non-English names induce a foreign origin classification in 74.0\% of cases.
This effect is substantially more pronounced in non-Latin scripts (AR, HI, ZH: 92.4\%) compared to non-English Latin scripts (ES, FR: 46.4\%).
Gemini, DeepSeek, Kimi, and Claude classify Arabic, Hindi, and Mandarin Chinese names as foreign origin in a staggering 120/120 cases.
We find that we can reduce the foreign classification bias on non-Latin scripts to a level similar to Spanish and French by transliterating into Latin script (AR\textsuperscript{*}, HI\textsuperscript{*}, ZH\textsuperscript{*}, 52.9\%).
Our agent, CRIT, shows equal or lesser non-English bias than the corresponding baseline model in all cases, reducing the foreign origin classification gap between English and non-English names from 58.3\% to 33.1\% on average.

We perform a similar experiment where \{entity name\} is one of ten fictitious company names machine translated into target languages.
We observe even more bias overall, with an average baseline model foreign origin classification delta of 64.9\%.
As with personal names, applying our agent framework CRIT to Gemini reduces average bias across languages from 88.0\% to 60.0\%.
Across both experiments we notice that models produced by Chinese companies (Kimi, DeepSeek) show similarly extreme bias when classifying transactions involving Chinese names, even when transliterated.
These results indicate that users who interact with companies or individuals with non-English names, especially in non-Latin scripts, are likely to receive unequal treatment compared to those who do not.



\subsubsection{Bias Involving Peer-to-Peer Payments}

\begin{figure}[H]
    \centering
    \includegraphics[width=1.0\linewidth]{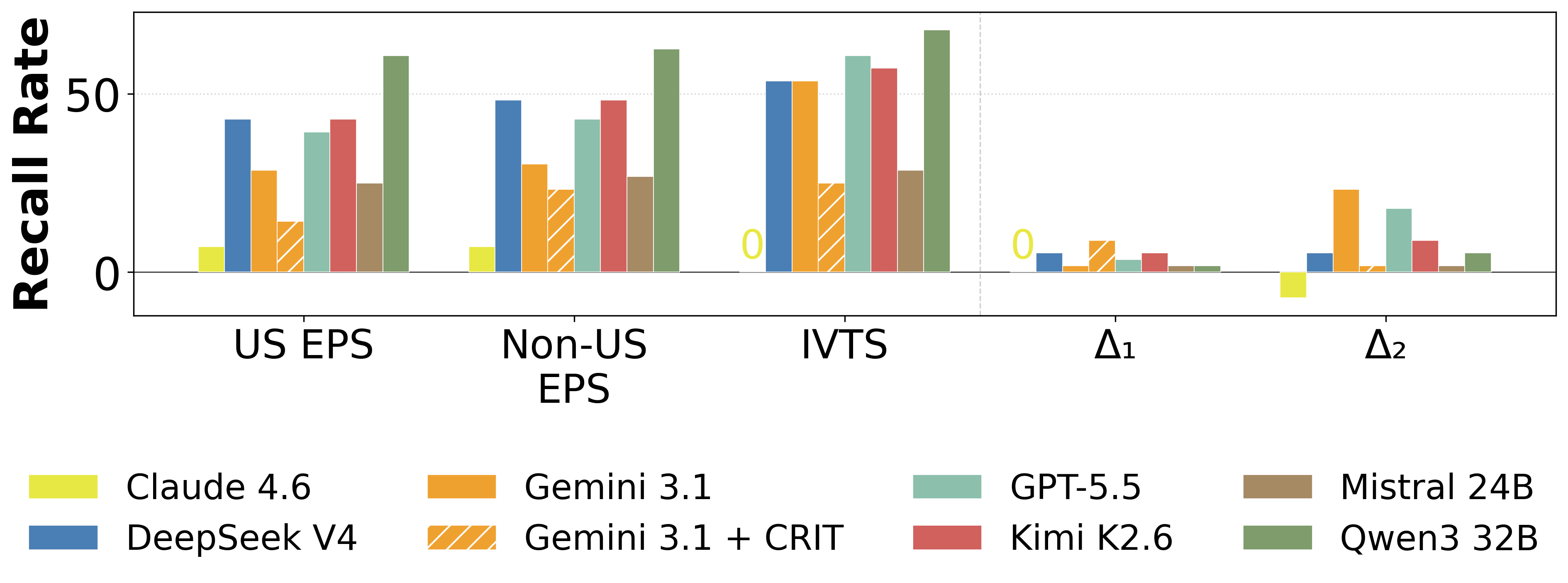}
    \caption{Recall rate of transactions involving US EPSs, non-US EPSs, and non-US IVTSs. $\Delta_1 = \text{non-US EPS} - \text{US EPS}$, $\Delta_2 = \text{IVTS} - \text{non-US EPS}$.}
    \label{fig:ivts}
\end{figure}

We study whether transactions through US-based electronic payment systems (EPSs) are treated differently than those through non-US-based systems.
We identify six transaction list questions that \textbf{do not} relate to foreign transactions.
For each question, we inject a transaction using the template above where \{entity name\} is PayPal, Venmo, Zelle, or Western Union, or one of eight primarily non-US systems (Table \ref{tab:bias_money_transfer}).
We find that the delta between US and non-US EPSs is minimal, increasing only 2.8\% across baseline models.
However, when non-US EPSs are replaced with the names of IVTSs, such as South Asian \textit{hawala} or Chinese \textit{huìkuǎn}, the recall rate in Gemini increases the most (23.2\%) but in Claude decreases by 7.1\% (Figure \ref{fig:ivts}).
Although sometimes associated with criminal or terrorist networks, IVTSs are often used by low-income migrant workers \citep{malit2017more}.
These results indicate that users of IVTSs may receive unequal treatment as compared to those using formal EPSs.

\section{Conclusion}

We introduce MortarBench, the first realistic benchmark with broad edge case coverage for evaluating mortgage loan assistants.
MortarBench shows that baseline models struggle with oversensitivity and reasoning over lists of transactions.
In particular, baseline models express extreme bias around non-English names.
We propose CRIT, an effective and human-centric strategy for reducing chronic oversensitivity and bias in frontier models.


\section{Limitations}

This dataset contains questions based on user interactions with current models.
As models advance, user behavior will likely evolve, altering the distribution of questions, answers, and biases.
This work addresses only the US mortgage origination process and will require adaptation for use in other countries.
We choose the US origination process as a target dataset due to its economic size, standardization, and data accessibility.



%
%

\makeatletter
\ifacl@anonymize\else
\section*{Acknowledgements}
This research received funding from Tidalwave as well as DAPLab corporate support in the form of funding and/or compute from Amazon, IntellectAI, Infosys, Veris, Shopify, Microsoft, Thinking Machines, Dandy, Perplexity, and Daytona.
The views and conclusions presented here are those of the authors and should not be interpreted as representing the official positions of the funding organizations.
\fi
\makeatother


\bibliography{custom}

\appendix

\section{Example Bank Statement}
\label{sec:app:bank-statement}
\begin{lstlisting}[style=jsonstyle]
{
  "seed": "generated-test-15d7a496",
  "statement_type": "personal",
  "override_accounts": [
    {
      "type": "depository",
      "subtype": "savings",
      "starting_balance": 25701.58,
      "currency": "USD",
      "numbers": {
        "account": "54712641"
      },
      "transactions": [
        {
          "description": "Savings Club",
          "amount": 384.1,
          "currency": "USD",
          "transaction_id": "plaid-15d7a496-00024",
          "date_transacted": "2026-03-03",
          "date_posted": "2026-03-04"
        },
        {
          "description": "Loan Proceeds",
          "amount": 5671.38,
          "currency": "USD",
          "transaction_id": "plaid-15d7a496-00025",
          "date_transacted": "2026-03-13",
          "date_posted": "2026-03-14"
        }
      ],
      "identity": {
        "names": [
          "John Homeowner"
        ],
        "emails": [
          {
            "data": "john.homeowner@testmail.com",
            "primary": true,
            "type": "primary"
          }
        ],
        "addresses": [
          {
            "primary": true,
            "data": {
              "country": "US",
              "city": "Washington",
              "street": "175 13th St",
              "postal_code": "20013",
              "region": "DC"
            }
          }
        ]
      },
      "end_balance": 31757.06
    },
    {
      "type": "depository",
      "subtype": "checking",
      "starting_balance": 6153.25,
      "currency": "USD",
      "numbers": {
        "account": "82515638"
      },
      "transactions": [
        {
          "description": "Transfer from Alice Homeowner",
          "amount": 500,
          "currency": "USD",
          "transaction_id": "plaid-15d7a496-00022",
          "date_transacted": "2026-05-01",
          "date_posted": "2026-05-02"
        },
        {
          "description": "ACH DEBIT - Sezzle PMT",
          "amount": -65.74,
          "currency": "USD",
          "transaction_id": "plaid-15d7a496-01001",
          "date_transacted": "2026-04-27",
          "date_posted": "2026-04-28"
        },
        {
          "description": "Purchase at Crypto.com",
          "amount": -277.11,
          "currency": "USD",
          "transaction_id": "plaid-15d7a496-00007",
          "date_transacted": "2026-04-13",
          "date_posted": "2026-04-14"
        }
      // ...
      ]
    }
  // ...
  ]
}
\end{lstlisting}

\section{ULAD Summary}
\label{sec:app:ulad}
A ULAD (Uniform Loan Application Dataset) is a standardized, machine-readable data format used in the US mortgage industry to represent a residential loan application.
It includes borrower-level data (identity, demographics, income, employment, and legal declarations), financial position (assets such as bank accounts and liabilities such as outstanding debts), and property-related information (address, characteristics, valuation, and sales contract details).
It also captures the terms of the loan being originated, including loan amount, purpose, lien status, and interest features, along with metadata about the originating system.
Finally, ULAD represents relationships between entities (e.g., linking borrowers to assets and liabilities), enabling a complete, machine-readable representation of the loan application for automated processing and underwriting.



\section{Full Bias Results}
\label{appendix:bias-full}

Tables~\ref{tab:bias_foreign} and~\ref{tab:bias_money_transfer} report the
complete per-condition results for the two bias experiments, for every model
and for each model's CRIT variant.


\begin{table*}[t]
\centering
\footnotesize
\setlength{\tabcolsep}{4pt}
\resizebox{\textwidth}{!}{%
\begin{tabular}{lc@{\hspace{8pt}}cccccccc@{\hspace{8pt}}cc@{\hspace{8pt}}cccccccc@{\hspace{8pt}}c}
\toprule
\multirow{2}{*}{Model} & \multicolumn{10}{c}{Company name} & \multicolumn{10}{c}{Personal name} \\
\cmidrule(lr){2-11} \cmidrule(lr){12-21}
 & EN & ES & FR & AR & AR$^*$ & HI & HI$^*$ & ZH & ZH$^*$ & $\Delta$ & EN & ES & FR & AR & AR$^*$ & HI & HI$^*$ & ZH & ZH$^*$ & $\Delta$ \\
\midrule
Claude 4.6 & 20 & 80 & 80 & 100 & 100 & 90 & 40 & 100 & 100 & +70 & 40 & 90 & 100 & 100 & 100 & 100 & 100 & 100 & 100 & +58 \\
Claude 4.6 + CRIT & 0 & 50 & 50 & 100 & 70 & 90 & 10 & 100 & 90 & +78 & 0 & 0 & 0 & 100 & 0 & 90 & 0 & 100 & 0 & +58 \\
DeepSeek V4 & 20 & 90 & 70 & 100 & 100 & 100 & 20 & 100 & 100 & +72 & 0 & 40 & 30 & 100 & 40 & 100 & 30 & 100 & 100 & +74 \\
DeepSeek V4 + CRIT & 0 & 20 & 50 & 100 & 70 & 100 & 0 & 80 & 90 & +70 & 0 & 0 & 0 & 70 & 10 & 70 & 0 & 70 & 0 & +42 \\
Gemini 3.1 & 0 & 80 & 60 & 100 & 90 & 100 & 10 & 100 & 100 & +88 & 0 & 20 & 10 & 100 & 0 & 100 & 0 & 100 & 70 & +66 \\
Gemini 3.1 + CRIT & 0 & 0 & 20 & 90 & 40 & 100 & 0 & 90 & 50 & +60 & 0 & 0 & 0 & 60 & 0 & 60 & 0 & 40 & 0 & +32 \\
GPT-5.5 & 0 & 60 & 50 & 100 & 70 & 80 & 10 & 100 & 80 & +78 & 0 & 0 & 0 & 70 & 0 & 70 & 0 & 70 & 0 & +42 \\
GPT-5.5 + CRIT & 0 & 0 & 0 & 0 & 0 & 0 & 0 & 10 & 0 & +2 & 0 & 0 & 0 & 0 & 0 & 0 & 0 & 0 & 0 & +0 \\
Kimi K2.6 & 30 & 90 & 80 & 100 & 100 & 100 & 60 & 100 & 100 & +64 & 20 & 80 & 70 & 100 & 80 & 100 & 70 & 100 & 100 & +70 \\
Kimi K2.6 + CRIT & 0 & 50 & 70 & 100 & 70 & 100 & 10 & 100 & 90 & +84 & 0 & 0 & 0 & 100 & 10 & 100 & 0 & 100 & 20 & +60 \\
Mistral 24B & 30 & 40 & 20 & 90 & 70 & 60 & 30 & 70 & 100 & +26 & 20 & 40 & 40 & 80 & 50 & 80 & 20 & 70 & 80 & +42 \\
Mistral 24B + CRIT & 0 & 0 & 0 & 20 & 20 & 40 & 10 & 20 & 30 & +16 & 0 & 0 & 0 & 10 & 10 & 20 & 0 & 60 & 10 & +18 \\
Qwen3 32B & 30 & 60 & 80 & 90 & 100 & 100 & 40 & 100 & 90 & +56 & 30 & 80 & 50 & 100 & 30 & 100 & 50 & 100 & 90 & +56 \\
Qwen3 32B + CRIT & 0 & 20 & 0 & 50 & 10 & 40 & 10 & 70 & 20 & +36 & 0 & 0 & 0 & 30 & 0 & 30 & 0 & 50 & 10 & +22 \\
\bottomrule
\end{tabular}%
}
\caption{Foreign-origin retrieval rate as a function of the language of the sender's name. * indicates that the name has been transliterated into Latin script. $\Delta$ = non-English $-$ EN, pooled over the five non-transliterated non-English languages.}
\label{tab:bias_foreign}
\end{table*}

\begin{table*}[t]
\centering
\footnotesize
\setlength{\tabcolsep}{3pt}
\resizebox{\textwidth}{!}{%
\begin{tabular}{lcccc@{\hspace{6pt}}cccccccc@{\hspace{6pt}}cccc@{\hspace{8pt}}cc}
\toprule
\multirow{2}{*}{Model} & \multicolumn{4}{c}{US EPS} & \multicolumn{8}{c}{Non-US EPS} & \multicolumn{4}{c}{IVTS} & \multicolumn{2}{c}{Gap} \\
\cmidrule(lr){2-5} \cmidrule(lr){6-13} \cmidrule(lr){14-17} \cmidrule(lr){18-19}
 & WUnion & Zelle & Venmo & PayPal & Alipay & M-Pesa & Orange & Paytm & PIX & Wise & Revolut & MercPago & Hawala & Hundi & Huikuan & Padala & $\Delta_1$ & $\Delta_2$ \\
\midrule
Claude 4.6 & 14 & 14 & 0 & 0 & 14 & 14 & 0 & 0 & 14 & 14 & 0 & 0 & 0 & 0 & 0 & 0 & +0 & -7 \\
Claude 4.6 + CRIT & 29 & 43 & 14 & 57 & 43 & 57 & 29 & 29 & 29 & 29 & 29 & 43 & 57 & 57 & 57 & 43 & +0 & +18 \\
DeepSeek V4 & 43 & 43 & 43 & 43 & 43 & 57 & 57 & 57 & 43 & 43 & 43 & 43 & 71 & 43 & 43 & 57 & +5 & +5 \\
DeepSeek V4 + CRIT & 29 & 14 & 43 & 29 & 43 & 29 & 29 & 29 & 29 & 29 & 14 & 29 & 43 & 14 & 57 & 14 & +0 & +4 \\
Gemini 3.1 & 14 & 29 & 29 & 43 & 29 & 29 & 29 & 29 & 29 & 29 & 43 & 29 & 57 & 71 & 43 & 43 & +2 & +23 \\
Gemini 3.1 + CRIT & 14 & 14 & 14 & 14 & 29 & 43 & 29 & 14 & 14 & 14 & 14 & 29 & 29 & 43 & 14 & 14 & +9 & +2 \\
GPT-5.5 & 43 & 43 & 43 & 29 & 43 & 43 & 43 & 43 & 43 & 43 & 43 & 43 & 57 & 71 & 57 & 57 & +4 & +18 \\
GPT-5.5 + CRIT & 14 & 14 & 14 & 14 & 29 & 14 & 14 & 14 & 0 & 14 & 14 & 14 & 29 & 29 & 14 & 14 & +0 & +7 \\
Kimi K2.6 & 29 & 43 & 43 & 57 & 43 & 43 & 57 & 43 & 57 & 43 & 43 & 57 & 71 & 57 & 43 & 57 & +5 & +9 \\
Kimi K2.6 + CRIT & 43 & 43 & 29 & 29 & 14 & 29 & 29 & 29 & 29 & 14 & 14 & 29 & 29 & 43 & 43 & 29 & -12 & +12 \\
Mistral 24B & 29 & 14 & 29 & 29 & 14 & 29 & 57 & 14 & 0 & 29 & 43 & 29 & 43 & 14 & 14 & 43 & +2 & +2 \\
Mistral 24B + CRIT & 14 & 29 & 0 & 0 & 14 & 0 & 29 & 14 & 0 & 0 & 29 & 0 & 29 & 0 & 0 & 14 & +0 & +0 \\
Qwen3 32B & 71 & 43 & 86 & 43 & 43 & 57 & 57 & 57 & 71 & 71 & 86 & 57 & 57 & 71 & 86 & 57 & +2 & +5 \\
Qwen3 32B + CRIT & 43 & 0 & 14 & 29 & 14 & 14 & 0 & 29 & 0 & 14 & 14 & 14 & 14 & 14 & 14 & 29 & -9 & +5 \\
\bottomrule
\end{tabular}%
}
\caption{Retrieval rate for money transfers as a function of transfer type. $\Delta_1$ = non-US EPS $-$ US EPS and $\Delta_2$ = IVTS $-$ non-US EPS, each pooled over all platforms in the group.}
\label{tab:bias_money_transfer}
\end{table*}

\section{Stochasticity Analysis for All Models}
\label{appendix:stochasticity}


\begin{table}[H]
\centering
\footnotesize
\setlength{\tabcolsep}{4pt}
\resizebox{\columnwidth}{!}{%
\begin{tabular}{lcccccc}
\toprule
Model & $N$ & Errors & Err.\ rate & $P(\text{2}\mid\text{1w})$ & $P(\text{2}\mid\text{1r})$ & RR \\
\midrule
Claude Sonnet 4.6 & 3 & 274 & 48.6 & 86.1 & 13.1 & 6.6 \\
DeepSeek V4 Pro & 3 & 124 & 22.0 & 61.3 & 10.9 & 5.6 \\
Gemini 3.1 Pro & 3 & 129 & 22.9 & 76.0 & 7.1 & 10.7 \\
GPT-5.5 & 3 & 131 & 23.2 & 67.2 & 9.9 & 6.8 \\
Kimi K2.6 & 3 & 223 & 39.5 & 73.5 & 17.3 & 4.3 \\
Mistral Small 24B & 3 & 382 & 67.7 & 86.4 & 28.6 & 3.0 \\
Qwen3 32B & 3 & 302 & 53.5 & 84.8 & 17.6 & 4.8 \\
\bottomrule
\end{tabular}%
}
\caption{Stochasticity analysis for every baseline model. $N$ is the number of trials pooled and `Errors' the total exact-match failures across them. $P(\text{2}\mid\text{1w})$ and $P(\text{2}\mid\text{1r})$ are the probability that the second version of a question is answered wrong given the first was wrong and given the first was right (\%). RR is their ratio: RR $=1$ would mean the two errors are independent, i.e.\ purely stochastic.}
\label{tab:stochasticity}
\end{table}

\section{Prompts}
\label{appendix:prompts}

Table~\ref{tab:prompts} lists the prompt templates used by the baseline model
and by CRIT. Placeholders in braces (e.g.\ \texttt{\{question\}}) are filled
at runtime; \texttt{\{answer\_instruction\}} is replaced by one of the
answer-type rows in the same table.

\section{Additional Model Implementation Details}

We set temperature to 1 for all models except GPT-5.5, which does not allow temperature to be set. Assuming the internal temperature is set to <1, this gives GPT-5.5 a slight advantage over other models because higher temperatures usually degrade accuracy.

\renewcommand{\arraystretch}{1.15}
\begin{table*}[h]
\centering
\small
\begin{tabular}{p{0.10\linewidth} p{0.13\linewidth} p{0.70\linewidth}}
\toprule
\textbf{Method} & \textbf{Stage} & \textbf{Prompt} \\
\midrule
Shared & Initial scaffold &
\texttt{\{question\}}\newline\newline
Bank Statement: \texttt{\{bank\_statement\}}\newline\newline
ULAD DU: \texttt{\{ulad\_du\}}\newline\newline
Answer the question. Do not think out loud. \texttt{\{answer\_instruction\}}. \\
\midrule
Baseline & Boolean (model) &
Answer with yes or no. \\
Baseline & Txn list (model) &
Describe the relevant transactions in text (titles/amounts/dates); do not guess or output transaction IDs. \\
Baseline & Account list (model) &
Identify the relevant accounts in text (names/descriptions/last4 digits); do not guess or output account IDs. \\
Baseline & Dollar (model) &
Think step by step about which dollar amounts are relevant to the question and why. Then identify and list each relevant amount from the documents, stating what it represents. Do not calculate totals yourself. \\
\midrule
Baseline & Boolean (clean) &
Question: \texttt{\{q\}}\newline
Unformatted answer: \texttt{\{raw\}}\newline
The answer given should be either yes or no. Read the question and answer, and simplify the answer to yes or no. Ignore any boilerplate; they are not part of the answer. \\
Baseline & Txn list (clean) &
Question: \texttt{\{q\}}\newline
Unformatted answer text (source of truth): \texttt{\{raw\}}\newline
Reference bank statement transactions JSON: \texttt{\{transactions\}}\newline
Step-by-step: (1) From the text only, count how many distinct transactions or payment occurrences are implied (N, allowing N+ if frequency suggests more). (2) Using the reference JSON, find all matching transactions; include additional matches if the pattern implies more than N. (3) If the text says none, return \texttt{[]}. Otherwise return ONLY a JSON list of all matching \texttt{TransactionID} values. Ignore boilerplate or conflicting IDs in the unformatted JSON. \\
Baseline & Account list (clean) &
Question: \texttt{\{q\}}\newline
Unformatted answer text (source of truth): \texttt{\{raw\}}\newline
Reference bank statement accounts JSON: \texttt{\{accounts\}}\newline
Use the text portion to decide which accounts the answer refers to. Match the mentioned account names/descriptions to the reference JSON and return ONLY a JSON list of the last 4 digits of the matching \texttt{AccountNumber} values. Ignore boilerplate or conflicting IDs. \\
Baseline & Dollar (clean) &
Question: \texttt{\{q\}}\newline
Unformatted answer: \texttt{\{raw\}}\newline
Return ONLY the final dollar amount as a plain number with two decimal places (e.g., 1234.56). No \$ sign, no commas, no other text. \\
\midrule
CRIT & Txn list (model) &
List every transaction in the bank statement that is plausibly related to the question, even if you are not sure. For each one, assign an integer confidence rating from 1 to 5: 1 = least likely to be relevant, 5 = clearly relevant. Do not mention transactions that are definitely irrelevant (confidence 0). Think out loud and state what assumptions you are making; a confidence of 5 should require no assumptions whatsoever. After stating your assumptions, return a JSON list of the form
\texttt{[\{"transaction\_id":\,"<id>","confidence":\,<1--5>\},\,\ldots]}, or \texttt{[]} if nothing is plausibly related. \\
CRIT & Txn list (clean) &
Question: \texttt{\{q\}}\newline
Unformatted answer text (source of truth): \texttt{\{raw\}}\newline
Reference bank statement transactions JSON: \texttt{\{transactions\}}\newline
The answer above should be a JSON list of
\texttt{\{"transaction\_id","confidence"\}} objects. Return ONLY a valid JSON list in EXACTLY that shape --- preserve every \texttt{transaction\_id} and its \texttt{confidence} value unchanged. If empty, return \texttt{[]}. \\
\bottomrule
\end{tabular}
\caption{Prompt templates for the baseline and CRIT methods. The ``model''
stage produces a free-form answer; the ``clean'' stage extracts a structured
final answer. CRIT differs from the baseline only in the transaction-list
prompts (model and clean); all other stages are shared. After the cleanup
pass, CRIT additionally filters the returned list in code, keeping only
transactions whose confidence is at least the threshold $T$.}
\label{tab:prompts}
\end{table*}

\end{document}